\documentclass[11pt,a4paper]{article}
\usepackage{authblk}
\usepackage[hyperref]{emnlp-ijcnlp-2019}
\usepackage{times}
\usepackage{latexsym}
\usepackage{graphicx}
\usepackage{float}
\usepackage{url}
\usepackage{color}
\usepackage{booktabs}
\usepackage{dblfloatfix}
\usepackage{amsmath}
\usepackage{bbm}
\usepackage{siunitx}
\usepackage{mathptmx}

\aclfinalcopy % Uncomment this line for the final submission

\newcommand{\norm}[1]{\left\lVert#1\right\rVert}
\newcommand{\negexp}[1]{\exp\left(-#1\right)}
\newcommand{\E}{\mathrm{E}}
\newcommand{\Var}{\mathrm{Var}}

\title{Text-based inference of moral sentiment change}

\author[*]{Jing Yi Xie}
\author[*]{Renato Ferreira Pinto, Jr.}
\author[ ]{Graeme Hirst}
\author[ ]{Yang Xu}
\affil[ ]{Department of Computer Science, University of Toronto, Toronto, Canada}
\affil[ ]{\texttt {jingyi.xie@mail.utoronto.ca, \{renato,gh,yangxu\}@cs.toronto.edu}}

\date{}

\begin{document}
\maketitle
\begin{abstract}
  We present a text-based framework for investigating moral sentiment change of the public via longitudinal corpora. Our framework is based on the premise that language use can inform people's moral perception toward right or wrong, and we build our methodology by exploring moral biases learned from diachronic word embeddings. We demonstrate how a parameter-free model supports inference of historical shifts in moral sentiment toward concepts such as slavery and democracy over centuries at three incremental levels: moral relevance, moral polarity, and fine-grained moral dimensions. We apply this methodology to visualizing moral time courses of individual concepts and analyzing the relations between psycholinguistic variables and rates of moral sentiment change at scale. Our work offers opportunities for applying natural language processing toward characterizing moral sentiment change in society.
\end{abstract}

\renewcommand*{\thefootnote}{\fnsymbol{footnote}}
\footnotetext[1]{Equal contribution.}
\renewcommand*{\thefootnote}{\arabic{footnote}}
\setcounter{footnote}{0}

%\section{Credits}

    \begin{figure*}[t!]
    \centering
    \includegraphics{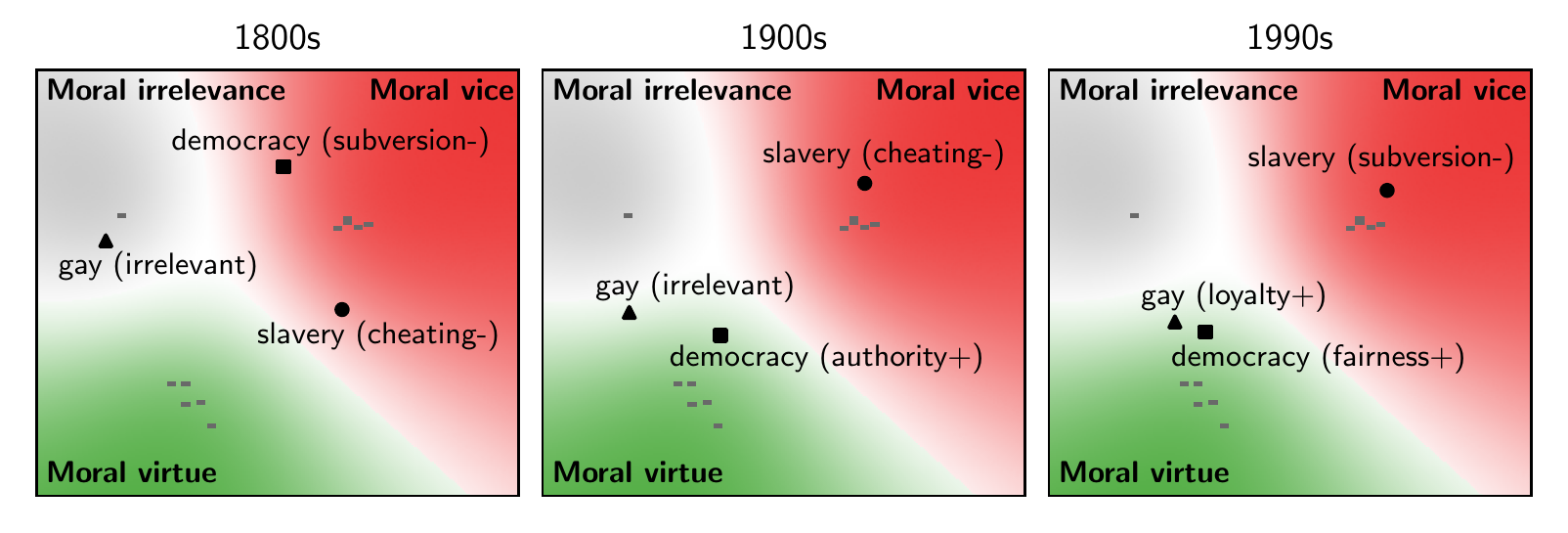}
    \caption{Illustration of moral sentiment change over the past two centuries. Moral sentiment trajectories of three probe concepts, {\it slavery}, {\it democracy}, and {\it gay}, are shown in moral sentiment embedding space through 2D projection from Fisher's discriminant analysis with respect to seed words from the classes of {\it moral virtue}, {\it moral vice}, and {\it moral irrelevance}. Parenthesized items represent moral categories predicted to be most strongly associated with the probe concepts. Gray markers represent the fine-grained centroids (or anchors) of these moral classes.}
    \label{fig:moral_map}
    \end{figure*}

\section{Moral sentiment change and language}

    People's moral sentiment---our feelings toward right or wrong---can change over time. For instance, the public's views toward {\it slavery} have shifted substantially over the past centuries \cite{oldfield2012popular}. How society's moral views evolve has been a long-standing issue and a constant source of controversy subject to  interpretations from social scientists, historians,  philosophers, among others. Here we ask whether natural language processing has the potential to inform moral sentiment change in society at scale, involving minimal human labour or intervention.
    
    %{\color{blue} We introduce various analyses that demonstrate our framework's ability to detect widespread attitudes. Our work is an automated approach to gaining implicit meaning from text and examining social phenomena, greatly aligning with many of the challenges present in NLP and computational social science.}
    % I've removed this for now due to space, but we can try to add a more concise version if fitting. [RF]

    The topic of moral sentiment has been thus far considered a traditional inquiry in philosophy \cite{hume1739tretise,smith1759,Kant2002}, with contemporary development of this topic represented in social psychology~\cite{piaget1932moral,kohlberg1969stage,stigler_cultural_1990,fiske1991social,pizarro_intelligence_2003}, cognitive linguistics~\cite{lakoff1996moral}, and more recently, the advent of Moral Foundations Theory~\cite{haidt2004intuitive,haidt2007moral,graham2013moral}. Despite the fundamental importance and interdisciplinarity of this topic, large-scale formal treatment of moral sentiment, particularly its evolution, is still in infancy from the natural language processing (NLP) community (see overview in Section~\ref{litreview}).

   We believe that there is a tremendous potential to bring NLP methodologies to bear on the problem of moral sentiment change. We build on extensive recent work showing that word embeddings reveal implicit human biases~\cite{bolukbasi2016man,caliskan2017semantics} and social stereotypes~\cite{garg2018word}. Differing from this existing work, we demonstrate that moral sentiment change can be revealed by moral biases implicitly learned from diachronic text corpora. Accordingly, we present to our knowledge the first text-based framework for probing moral sentiment change at a large scale with support for different levels of analysis concerning {\it moral relevance}, {\it moral polarity}, and {\it fine-grained moral dimensions}. As such, for any query item such as {\it slavery}, our goal is to automatically infer its moral trajectories from sentiments at each of these levels over a long  period of time.
   
  Our approach is based on the premise that people's moral sentiments are reflected in natural language, and more specifically, in text \cite{bloom2010morals}. In particular, we know that books are highly effective tools for conveying moral views to the public. For example, \textit{Uncle Tom's Cabin} \cite{stowe1852uncle} was central to the anti-slavery movement in the United States. The framework that we develop builds on this premise to explore changes in moral sentiment reflected in longitudinal or historical text.
   
   Figure~\ref{fig:moral_map} offers a preview of our framework by visualizing the evolution trajectories of the public's moral sentiment toward concepts signified by the probe words \textit{slavery}, \textit{democracy}, and \textit{gay}. Each of these concepts illustrates a piece of ``moral history'' tracked through a period of 200 years (1800 to 2000), and our framework is able to capture nuanced moral changes. For instance, {\it slavery} initially lies at the border of moral virtue (positive sentiment) and vice (negative sentiment) in the 1800s yet gradually moves toward the center of moral vice over the 200-year period; in contrast, {\it democracy} considered morally negative (e.g., subversion and anti-authority under monarchy) in the 1800s is now perceived as morally positive, as a mechanism for fairness; \textit{gay}, which came to denote homosexuality only in the 1930s~\cite{historical_thesaurus_english}, is inferred to be morally irrelevant until the modern day. We will describe systematic evaluations and applications of our framework that extend beyond these anecdotal cases of moral sentiment change.
   
   %Our framework captures the nuanced character of moral change, whereby words can undergo change by becoming morally relevant (e.g., \textit{gay} came to denote a sexual orientation in the 1930s; \citeauthor{historical_thesaurus_english}, 2019); changing toward positive (\textit{democracy}) or negative (\textit{slavery}) moral polarity; or in the fine-grained moral categories associated with a concept (e.g., care/harm or authority/subversion). We propose a multi-tiered framework that detects moral sentiment change at different  levels of granularity.
   
   The general text-based framework that we propose consists of a parameter-free approach that facilitates the prediction of public moral sentiment toward individual concepts, automated retrieval of morally changing concepts, and broad-scale psycholinguistic analyses of historical rates of moral sentiment change. We provide a description of the probabilistic models and data used, followed by comprehensive evaluations of our methodology.
	
\section{Emerging NLP research on morality} \label{litreview}
    
    An emerging body of work in natural language processing and computational social science has investigated how NLP systems can detect moral sentiment in online text. For example, moral rhetoric in social media and political discourse \cite{garten2016morality,johnson2018classification, lin2018acquiring}, the relation between moralization in social media and violent protests \cite{mooijman2018moralization}, and bias toward refugees in talk radio shows \cite{gillani2019simple} have been some of the topics explored in this line of inquiry. In contrast to this line of research, the development of a formal framework for moral sentiment change is still under-explored, with no existing systematic and formal treatment of this topic \cite{bloom2010morals}.
    
    While there is emerging awareness of ethical issues in NLP \cite{hovy2017proceedings,alfano2018proceedings}, work exploiting NLP techniques to study principles of moral sentiment change is scarce.  Moreover, since morality is variable across cultures and time \cite{graham2013moral,bloom2010morals}, developing systems that capture the diachronic nature of moral sentiment will be a pivotal research direction. Our work leverages and complements existing research that finds implicit human biases from word embeddings~\cite{bolukbasi2016man,caliskan2017semantics,garten2016morality} by developing a novel perspective on using NLP methodology to discover principles of moral sentiment change in human society.
    
    %Thus, inferring morally relevant topics and moral sentiment change at scale remains an open area of ethical studies in NLP.
    
    %In this work, we present a formal framework for large-scale inference of diachronic moral sentiment change from text, paving the way toward diachronic models of morality in NLP.

\section{A three-tier modelling framework}

    \begin{table*}[tbp]
        \centering
        \begin{tabular}{l c r @{\hspace*{1mm}} l}
        \toprule
        \bfseries Model & \bfseries Parameter & \multicolumn{2}{c}{\bfseries Posterior Inference}  \\
        \midrule
        Centroid          & --       &  $p(c\,|\,\mathbf{q}) \propto$ & $ \negexp{\norm{\mathbf{q}-\E\left[ \mathbf{S}_c \right]}} $ \\
        %\midrule
        Na\"ive Bayes     & --       &  $p(c\,|\,\mathbf{q}) \propto$ & $ \prod_{j=1}^d f_{N}\left(\mathbf{q}_j ; \mu=\E\left[\mathbf{S}_{c,j}\right], \sigma^2=\Var\left[\mathbf{S}_{c,j}\right]\right) $   \\
        %\midrule
        $k$-Nearest Neighbors ($k$NN)             & $k$      &  $p(c\,|\,\mathbf{q}) \propto$ & $ \big\lvert \left\{ \text{$k$ nearest seed words to $\mathbf{q}$} \right\} \cap \mathbf{S}_c \big\rvert $    \\
        %\midrule
        Kernel Density Estimation (KDE)              & $h$      &  $p(c\,|\,\mathbf{q}) \propto$ & $\frac{1}{\lvert \mathbf{S}_c \rvert} \sum_{\mathbf{w} \in \mathbf{S}_c} f_{MN}\left(\mathbf{q}; \mu = \mathbf{w}, \Sigma = \mathrm{diag}(h) \right)$     \\
        %\bottomrule
        \end{tabular}
        
        \caption{Summary of models for moral sentiment classification. Each model infers moral sentiment of a query word vector $\mathbf{q}$  based on moral classes $c$  (at any of the three levels) represented by moral seed words $\mathbf{S}_c$.  $\E\left[\mathbf{S}_c\right]$ is the mean vector of $\mathbf{S}_c$; $\E\left[\mathbf{S}_{c,j}\right], \Var\left[\mathbf{S}_{c,j}\right]$ refer to the mean and variance of $\mathbf{S}_c$ along the $j$-th  dimension in embedding space. $d$ is the number of embedding dimensions; and $f_N, f_{MN}$ refer to the density functions of univariate and multivariate normal distributions, respectively.}
        
        %\caption{Models caption, explain notation {\color{blue}I find these equations overloaded --- can these be simplified somehow, e.g. reducing symbols? The critical probabilsitic formulation is missing, e.g. what is your input and output variables?}}
        \label{tab:model_formulations}
    \end{table*}
    
    \begin{figure}
        \centering
        \includegraphics[width=\columnwidth]{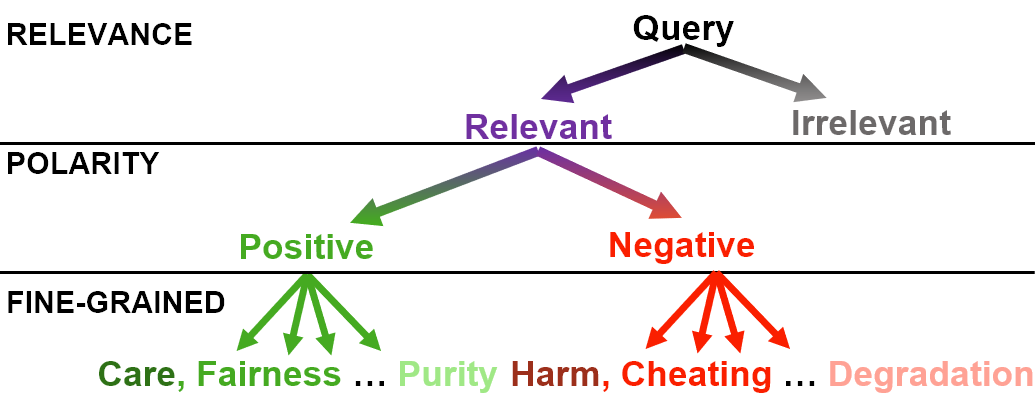}
        \caption{Illustration of the three-tier framework that supports moral sentiment inference at different levels.}
        \label{fig:tiers}
    \end{figure}
	Our framework treats the moral sentiment toward a concept at three incremental levels, as illustrated in Figure~\ref{fig:tiers}. First, we consider moral relevance, distinguishing between morally irrelevant and morally relevant concepts. At the second tier, moral polarity, we further split morally relevant concepts into those that are positively or negatively perceived in the moral domain. Finally, a third tier classifies these concepts into fine-grained categories of human morality.
	
    We draw from research in social psychology to inform our methodology, most prominently Moral Foundations Theory \cite[MFT;][]{graham2013moral}. MFT seeks to explain the structure and variation of human morality across cultures, and proposes five moral foundations: Care~/ Harm, Fairness~/ Cheating, Loyalty~/ Betrayal, Authority~/ Subversion, and Sanctity~/ Degradation. Each foundation is summarized by a positive and a negative pole, resulting in ten fine-grained moral categories.
    
    \subsection{Lexical data for moral sentiment}
    To ground moral sentiment in text, we leverage the Moral Foundations Dictionary \cite[MFD;][]{graham2009liberals}. The MFD is a psycholinguistic resource that associates each MFT category with a set of \textit{seed words}, which are words that provide evidence for the corresponding moral category in text. We use the MFD for moral polarity classification by dividing seed words into positive and negative sets, and for fine-grained categorization by splitting them into the 10 MFT categories.
    
    To implement the first tier of our framework and detect moral relevance, we complement our morally relevant seed words with a corresponding set of seed words approximating moral irrelevance based on the notion of valence, i.e., the degree of pleasantness or unpleasantness of a stimulus. We refer to the emotional valence ratings collected by \citet{warriner2013norms} for approximately 14,000 English words, and choose the words with most neutral valence rating that do not occur in the MFD as our set of morally irrelevant seed words, for an equal total number of morally relevant and morally irrelevant words.

    \subsection{Models}
    
    We propose and evaluate a set of probabilistic models to classify concepts in the three tiers of morality specified above. Our models exploit the semantic structure of word embeddings \cite{mikolov2013distributed} to perform tiered moral classification of query concepts. In each tier, the model receives a query word embedding vector $\mathbf{q}$ and a set of seed words for each class in that tier, and infers the posterior probabilities over the set of classes $c$ to which the query concept is associated with.
    
    The seed words function as ``labelled examples'' that guide the moral classification of novel concepts, and are organized per classification tier as follows. In moral relevance classification, sets $\mathbf{S}_0$ and $\mathbf{S}_1$ contain the morally irrelevant and morally relevant seed words, respectively; for moral polarity, $\mathbf{S}_+$ and $\mathbf{S}_-$ contain the positive and negative seed words; and for fine-grained moral categories, $\mathbf{S}_1, \ldots, \mathbf{S}_{10}$ contain the seed words for the 10 categories of MFT. Then our general problem is to estimate $p(c\,|\,\mathbf{q})$, where $\mathbf{q}$ is a query vector and $c$ is a moral category in the desired tier.
    
    We evaluate the following four models:
    \begin{itemize}
        \item A Centroid model summarizes each set of seed words by its expected vector in embedding space, and classifies concepts into the class of closest expected embedding in Euclidean distance following a softmax rule;
        \item A Na\"ive Bayes model considers both mean and variance, under the assumption of independence among embedding dimensions, by fitting a normal distribution with mean vector and diagonal covariance matrix to the set of seed words of each class;
        \item A $k$-Nearest Neighbors ($k$NN) model exploits local density estimation and classifies concepts according to the majority vote of the $k$ seed words closest to the query vector;
        \item A Kernel Density Estimation (KDE) model performs density estimation at a broader scale by considering the contribution of each seed word toward the total likelihood of each class, regulated by a bandwidth parameter $h$ that controls the sensitivity of the model to distance in embedding space.
    \end{itemize}
    Table~\ref{tab:model_formulations} specifies the formulation of each model. Note that we adopt a parsimonious design principle in our modelling: both Centroid and Na\"ive Bayes are parameter-free models, $k$NN only depends on the choice of $k$, and KDE uses a single bandwidth parameter $h$.

\section{Historical corpus data}

    To apply our models diachronically, we require a word embedding space that captures the meanings of words at different points in time and reflects changes pertaining to a particular word as diachronic shifts in a common embedding space.
	
	Following \citet{hamilton2016diachronic}, we combine skip-gram word embeddings \cite{mikolov2013distributed} trained on longitudinal corpora of English with rotational alignments of embedding spaces to obtain diachronic word embeddings that are aligned through time.
	
	We divide historical time into decade-long bins, and use two sets of embeddings provided by \citet{hamilton2016diachronic}, each trained on a different historical corpus of English:
	
	\begin{itemize}
	    \item Google N-grams \cite{lin2012syntactic}: a corpus of $8.5 \times 10^{11}$ tokens  collected from the English literature (Google Books, all-genres) spanning the period 1800--1999.
	    
	    \item COHA \cite{davies2010coha}: a smaller corpus of $4.1 \times 10^8$ tokens from works selected so as to be genre-balanced and representative of American English in the period 1810--2009.
	\end{itemize}

\section{Model evaluations}

	We evaluated our models in two ways: classification of moral seed words on all three tiers (moral relevance, polarity, and fine-grained categories), and correlation of model predictions with human judgments. 
	
	\subsection{Moral sentiment inference of seed words}
	
	In this evaluation, we assessed the ability of our models to classify the seed words that compose our moral environment in a leave-one-out classification task. We performed the evaluation for all three classification tiers: 1) moral relevance, where seed words are split  into morally relevant and morally irrelevant; 2) moral polarity, where moral seed words are split into positive and negative; 3) fine-grained categories, where moral seed words are split into the 10 MFT categories. In each test, we removed one seed word from the training set at a time to obtain cross-validated model predictions.
	
	Table~\ref{tab:model_comparison} shows classification accuracy for all models and corpora on each tier for the 1990--1999 period.\footnote{We also computed average accuracy over all decades using Google N-grams, the only corpus covering all moral categories through time. See Supplementary Material.}
	We observe that all models perform substantially better than chance, confirming the efficacy of our methodology in capturing moral dimensions of words. We also observe that models using word embeddings trained on Google N-grams perform better than those trained on COHA, which could be expected given the larger corpus size of the former.
	
	In the remaining analyses, we employ the Centroid model, which offers competitive accuracy and a simple, parameter-free specification.

    \begin{table*}[tbp]
    \centering
\begin{tabular}{lcccccc}
\toprule              
                    &      \multicolumn{3}{c}{Google N-grams}                                  &       \multicolumn{3}{c}{COHA} \\
\cmidrule(lr){2-4}
\cmidrule(lr){5-7}
    \bfseries Model &  \bfseries  Relevance  &  \bfseries Polarity &  \bfseries Category     &     \bfseries  Relevance  &  \bfseries Polarity &  \bfseries Category \\                           
\midrule                                            
    Random &                        0.50 &               0.50 &                0.10      &                   0.50 &                  0.50 &                0.10   \\       
 Centroid &              \bfseries 0.84 &                0.90 &                \bfseries 0.59      &                   0.78 &        \bfseries 0.80 &      \bfseries 0.40            \\          
       Na\"ive Bayes &        \bfseries 0.84 &                0.89 &                0.53        &                 0.76 &                  0.78 &                0.39              \\       
    1-NN &                         0.80 &                0.88 &                0.53       &                  0.74 &                  0.76 &                0.32           \\           
    5-NN &                         0.83 &      \bfseries 0.93 &                0.57      &                   0.74 &                  0.75 &                0.33          \\           
      KDE &                        0.82 &                0.90 &                0.57     &                   \bfseries 0.80 &                  0.76 &                0.33             \\                            
%\bottomrule                                                                        
\end{tabular} 
    \caption{Classification accuracy of moral seed words for moral relevance, moral polarity, and fine-grained moral categories based on 1990--1999 word embeddings for two independent corpora, Google N-grams and COHA.}
    \label{tab:model_comparison}
\end{table*}

\subsection{Alignment with human valence ratings}
    
    \begin{table}[tbp]
    \centering
    \begin{tabular}{ll}
    {\bf Corpus} & {\bf Correlation} \\\hline
    Google N-grams & 0.43 ($n=12293; p< 0.0001$) \\
    COHA & 0.38 ($n=7141; p< 0.0001$)
    \end{tabular}
    \caption{Pearson correlations between model predicted moral sentiment polarities and human valence ratings.}\label{tab:valence}
    \end{table}
    
    We evaluated the approximate agreement between our methodology and human judgments using valence ratings, i.e., the degree of pleasantness or unpleasantness of a stimulus. Our assumption is that the valence of a concept should correlate with its perceived moral polarity, e.g., morally repulsive ideas should evoke an unpleasant feeling. However, we do not expect this correspondence to be perfect; for example, the concept of \textit{dessert} evokes a pleasant reaction without being morally relevant.
    
    In this analysis, we took the valence ratings for the nearly 14,000 English nouns collected by \citet{warriner2013norms} and, for each query word $q$, we generated a corresponding prediction of positive moral polarity from our model, $P(c_+\,|\,\mathbf{q})$. Table~\ref{tab:valence} shows the correlations between human valence ratings and predictions of positive moral polarity generated by models trained on each of our corpora. We observe that the correlations are significant, suggesting the ability of our methodology to capture relevant features of moral sentiment from text.
    
    In the remaining applications, we use the diachronic embeddings trained on the Google N-grams corpus, which enabled superior model performance throughout our evaluations.

\begin{table*}[!ht]
\centering
\begin{tabular}{@{}lScc@{}}\hline
% \begin{tabular}{c|c|c|c}\hline
\textbf{Concept} & \textbf{Rate (e$-$03/decade)} & \textbf{Relevant Moral Category} & \textbf{Switching Period}   \\\hline
abortion & 4.24** & cheating$-$ & 1890\\
propriety & 4.06*** & fairness+ & 1870\\
commandment & 3.68*** & sanctity+ & 1880\\
righteousness & 3.56*** & sanctity+ & 1800\\
authorities & 3.42*** & authority+ & 1890\\
apostle & 3.41*** & authority+ & 1900\\
intervention & 3.33** & authority+ & 1860\\
jew & 3.32*** & degradation$-$ & 1870\\
foreigner & 3.30*** & authority+ & 1800\\
individuality & 3.26*** & authority+ & 1860\\
\end{tabular}
\caption{Top 10 changing words towards moral relevance during 1800--2000, with model-inferred moral category and switching period. *, **, and *** denote $p<0.05$, $p<0.001$, and $p< 0.0001$, all Bonferroni-corrected.}
\label{tab:relevance_changing}
\end{table*}

\begin{table*}[!ht]
\centering
\begin{tabular}{@{}lSccc@{}}\hline
\textbf{Concept} & \textbf{Rate (e$-$03/decade)} & \textbf{Early Category} & \textbf{Modern Category} & \textbf{Switching Period}   \\\hline
wage & 4.38** & subversion$-$ & fairness+ & 1810\\
commitment & 3.78 & harm$-$ & authority+ & 1830\\
innovation & 3.21 & authority+ & authority+ & 1800\\
help & 3.20** & sanctity+ & care+ & 1880\\
mandate & 3.17* & authority+ & authority+ & 1800\\
guidance & 3.04*** & authority+ & authority+ & 1800\\
licence & 3.01* & authority+ & authority+ & 1800\\
abortion & 2.95 & degradation$-$ & cheating$-$ & 1890\\
democracy & 2.93** & authority+ & fairness+ & 1800\\
disclosure & 2.89*** & fairness+ & fairness+ & 1840\\\hline
propaganda & -7.75* & authority+ & subversion$-$ & 1910\\
humiliation & -4.05*** & authority+ & harm$-$ & 1800\\
seriousness & -4.02** & authority+ & harm$-$ & 1800\\
legacy & -3.76*** & authority+ & subversion$-$ & 1800\\
behavior & -3.73** & harm$-$ & authority+ & 1830\\
cheerfulness & -3.66*** & authority+ & sanctity+ & 1800\\
candour & -3.37* & authority+ & degradation$-$ & 1800\\
offense & -3.37 & fairness+ & subversion$-$ & 1840\\
indulgence & -3.37*** & authority+ & degradation$-$ & 1800\\
exertion & -3.26*** & authority+ & degradation$-$ & 1800\\
\end{tabular}
\caption{Top 10 changing words towards moral positive (upper panel) and negative (lower panel) polarities, with model-inferred most representative moral categories during historical and modern periods and the switching periods. *, **, and *** denote $p<0.05$, $p<0.001$, and $p<0.0001$, all Bonferroni-corrected for multiple tests.}
\label{tab:polarity_changing}
\end{table*}

\section{Applications to  diachronic morality}

    We applied our framework in three ways: 1) evaluation of selected concepts in historical time courses and prediction of human judgments; 2) automatic detection of moral sentiment change; and 3) broad-scale study of the relations between psycholinguistic variables and historical change of moral sentiment toward concepts.
    
   \subsection{Moral change in individual concepts}
\begin{figure*}[!ht]
    \centering
    \includegraphics[width=0.495\textwidth]{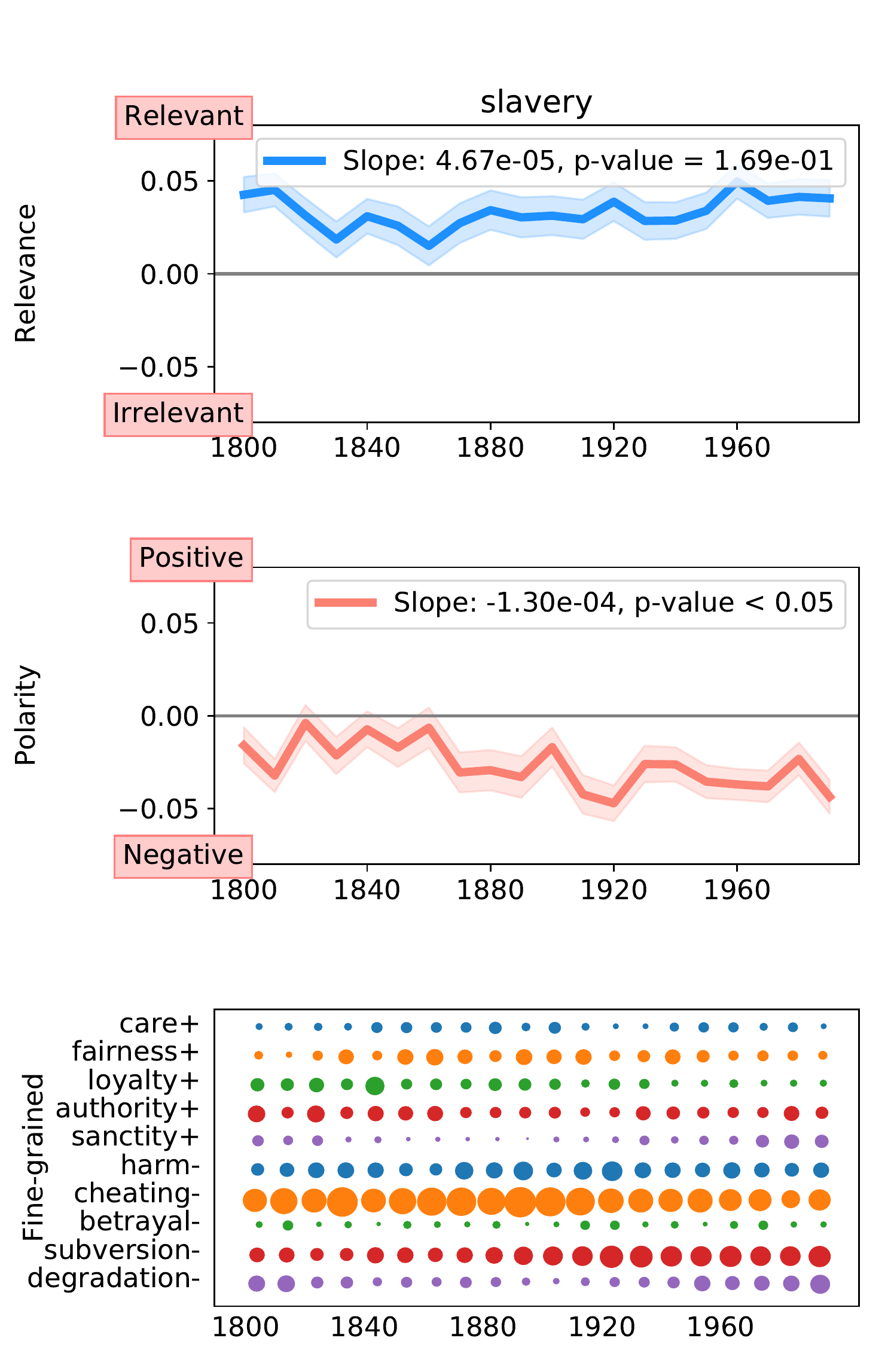}
    \includegraphics[width=0.495\textwidth]{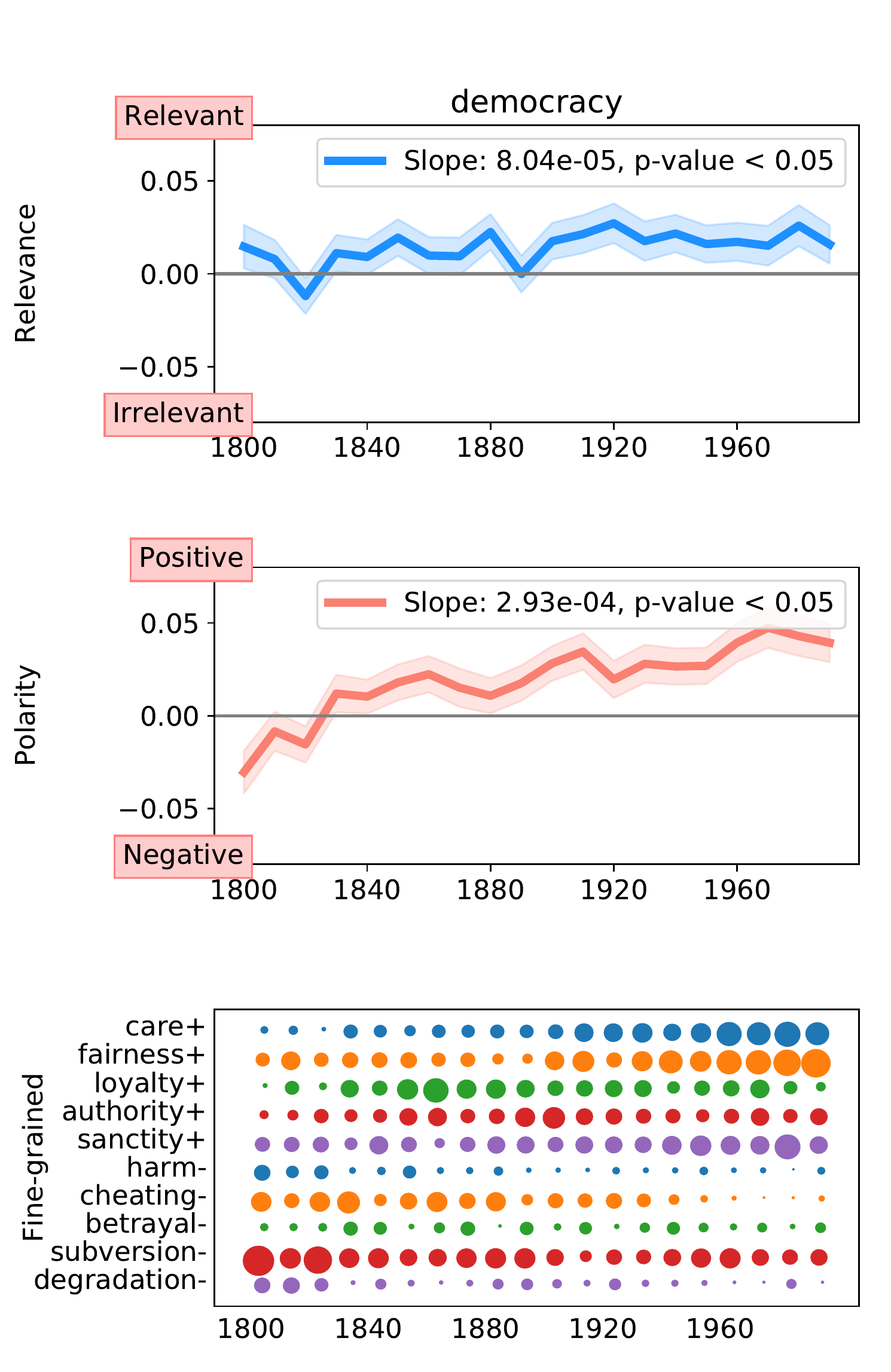}
    \caption{Moral sentiment time courses of {\it slavery} (left) and {\it democracy} (right) at each of the three levels, inferred by the Centroid model. Time courses at the moral relevance and polarity levels are in log odds ratios, and those for the fine-grained moral categories are represented by circles with sizes proportional to category probabilities.}
    \label{fig:time_courses}
    \end{figure*}

   \paragraph{Historical time courses.}
    We applied our models diachronically to predict time courses of moral relevance, moral polarity, and fine-grained moral categories toward two historically relevant topics: slavery and democracy. By grounding our model in word embeddings for each decade and querying concepts at the three tiers of classification, we obtained the time courses shown in Figure~\ref{fig:time_courses}.
    
    We note that these trajectories illustrate actual historical trends. Predictions for democracy show a trend toward morally positive sentiment, consistent with the adoption of democratic regimes in Western societies. On the other hand, predictions for slavery trend down and suggest a drop around the 1860s, coinciding with the American Civil War. We also observe changes in the dominant fine-grained moral categories, such as the perception of democracy as a fair concept, suggesting potential mechanisms behind the polarity changes and providing further insight into the public sentiment toward these concepts as evidenced by text.
    
    \paragraph{Prediction of human judgments.}
    We explored the predictive potential of our framework by comparing model predictions with human judgments of moral relevance and acceptability. We used data from the Pew Research Center's 2013 Global Attitudes survey \cite{nw_global_2014}, in which participants from 40 countries judged 8 topics such as \textit{abortion} and \textit{homosexuality} as one of ``acceptable", ``unacceptable", and ``not a moral issue".
    
    We compared human ratings with model predictions at two tiers: for moral relevance, we paired the proportion of ``not a moral issue'' human responses with irrelevance predictions $p(c_0\,|\,\mathbf{q})$ for each topic, and for moral acceptability, we paired the proportion of ``acceptable'' responses with positive predictions $p(c_+\,|\,\mathbf{q})$. We used 1990s word embeddings, and obtained predictions for two-word topics by querying the model with their averaged embeddings. 
    
\begin{figure*}[!ht]
	    \centering
	    \includegraphics[width=\columnwidth]{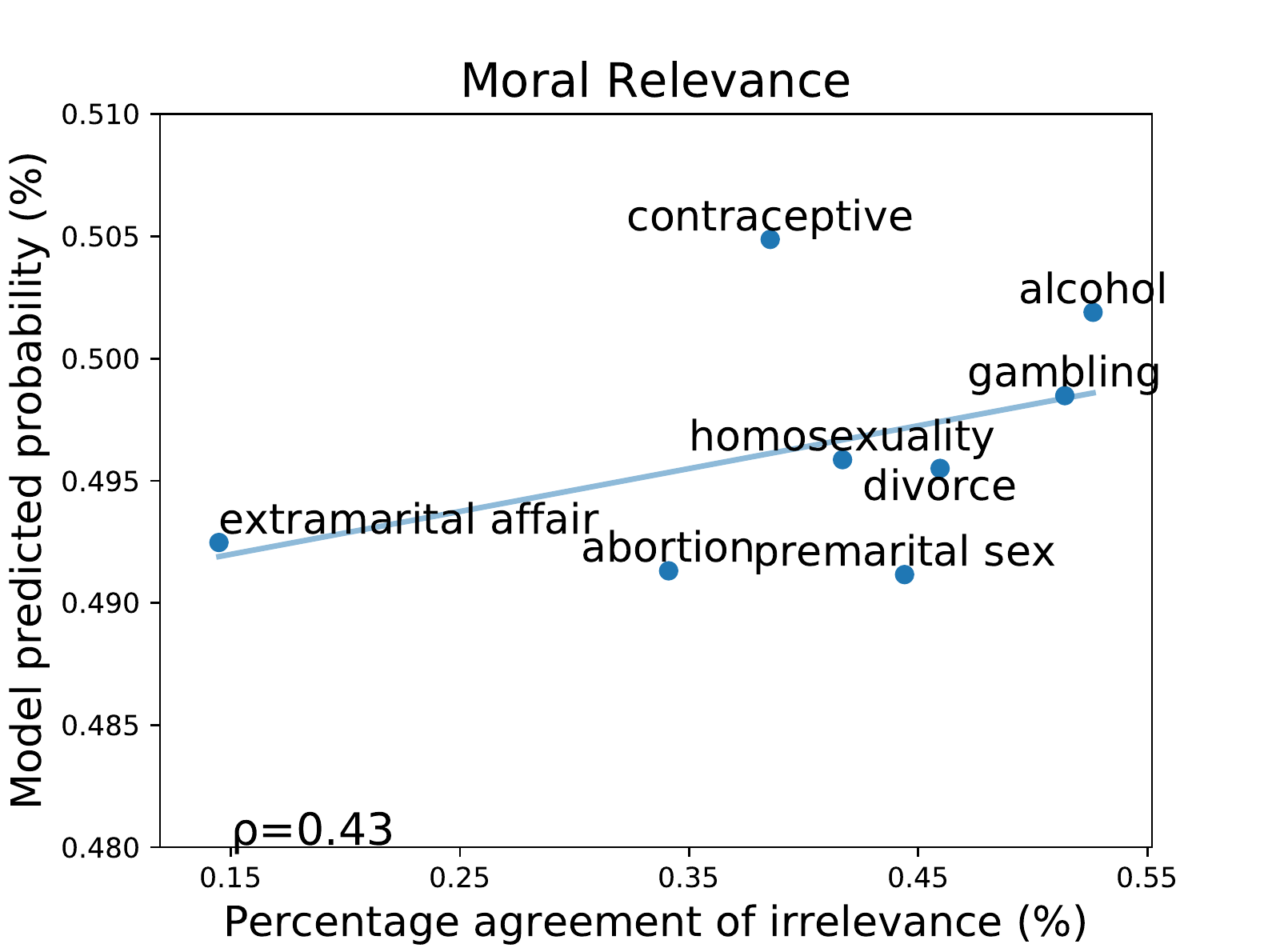}
	    \includegraphics[width=\columnwidth]{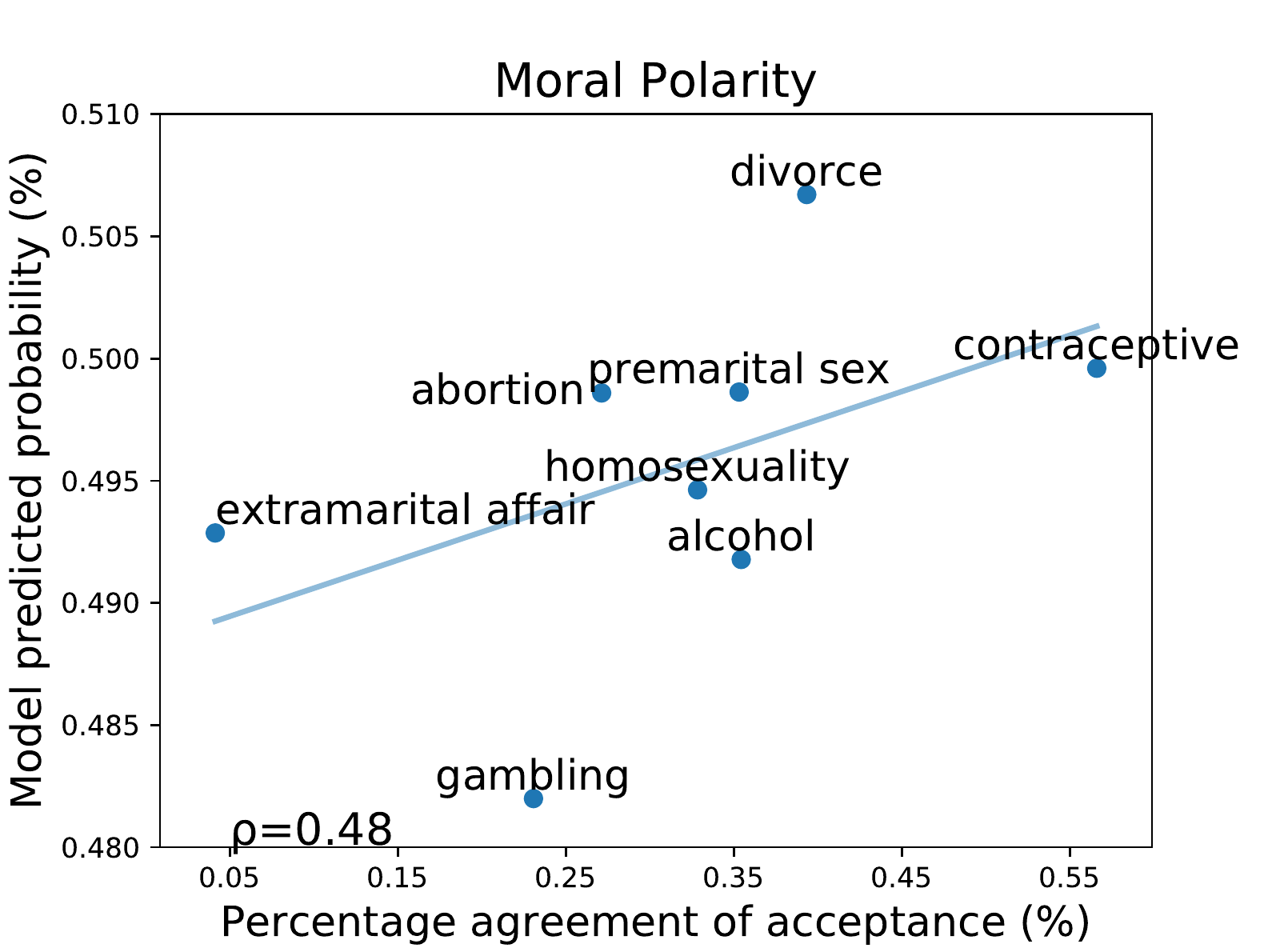}
	    \caption{Model predictions against percentage of Pew respondents who selected ``Not a moral concern" (left) or ``Acceptable" (right), with lines of best fit and Pearson correlation coefficients $\rho$ shown in the background.}
	    \label{fig:pew}
	\end{figure*}

    Figure~\ref{fig:pew} shows plots of relevance and polarity predictions against survey proportions, and we observe a visible correspondence between model predictions and human judgments despite the difficulty of this task and limited number of topics.
    
    \subsection{Retrieval of morally changing concepts}
    
    Beyond analyzing selected concepts, we applied our framework predictively on a large repertoire of words to automatically discover the concepts that have exhibited the greatest change in moral sentiment at two tiers, moral relevance and moral polarity.
    
    We selected the 10,000 nouns with highest total frequency in the 1800--1999 period according to data from \citet{hamilton2016diachronic}, restricted to words labelled as nouns in WordNet \cite{miller1995wordnet} for validation. For each such word $\mathbf{q}$, we computed diachronic moral relevance scores $R_i = p(c_1\,|\,\mathbf{q}), i=1,\ldots,20$ for the 20 decades in our time span. Then, we performed a linear regression of $R$ on $T = 1,\ldots,n$ and took the fitted slope as a measure of  moral relevance change. We repeated the same procedure for moral polarity. Finally, we removed words with average relevance score below $0.5$ to focus on morally relevant retrievals.
    
    Table~\ref{tab:relevance_changing} shows the words with steepest predicted change toward moral relevance, along with their predicted fine-grained moral categories in modern times (i.e., 1900--1999). Table~\ref{tab:polarity_changing} shows the words with steepest predicted change toward the positive and negative moral poles. To further investigate the moral sentiment that may have led to such polarity shifts, we also show the predicted fine-grained moral categories of each word at its earliest time of predicted moral relevance and in modern times. Although we do not have access to ground truth for this application, these results offer initial insight into the historical moral landscape of the English language at scale.
    
	\subsection{Broad-scale investigation of moral change}
	
	In this application, we investigated the hypothesis that concept concreteness is inversely related to change in moral relevance, i.e., that concepts considered more abstract might become morally relevant at a higher rate than concepts considered more concrete. To test this hypothesis, we performed a multiple linear regression analysis on rate of change toward moral relevance of a large repertoire of words against concept concreteness ratings, word frequency \cite[a correlate of semantic change; see][]{hamilton2016diachronic}, and word length \cite[as a proxy for concept complexity; see][]{lewis2016length}.
	
	We obtained norms of concreteness ratings from \citet{warriner2013norms}. We collected the same set of high-frequency nouns as in the previous analysis, along with their fitted slopes of moral relevance change. Since we were interested in moral relevance change within this large set of words, we restricted our analysis to those words whose model predictions indicate change in moral relevance, in either direction, from the 1800s to the 1990s.
	
	We performed a multiple linear regression under the following model:
		\begin{equation}
	    \label{eq:regression}
	    \Tilde{\rho}(w) = \beta_f \log(f(w)) + \beta_l l(w) + \beta_c c(w) + \beta_0 + \Tilde{\epsilon}
	\end{equation}
		Here $\rho(w)$ is the slope of moral relevance change for word $w$; $f(w$) is its average frequency; $l(w)$ is its character length; $c(w)$ is its concreteness rating; $\beta_f$, $\beta_l$, $\beta_c$, and $\beta_0$ are the corresponding factor weights and intercept, respectively; and $\epsilon \sim \mathcal{N}(0, \sigma)$ is the regression error term.
	
	Table~\ref{tab:regression} shows the results of multiple linear regression. We observe that concreteness is a significant negative predictor of change toward moral relevance, suggesting that abstract concepts are more strongly associated with increasing moral relevance over time than concrete concepts. This significance persists under partial correlation test against the control factors ($p < 0.01$).
	
	\begin{table}[tb]
	    \centering
	    \begin{tabular}{lS[table-format=1.2]S[table-format=1.3]}
	         \toprule
	         \bfseries Factor & \bfseries Coeff. (e-03) &  \bfseries Significance \\
	         \midrule
	          Frequency       & 0.1      & {$p<0.001$} \\ 
	         Length                    & -0.03    & {n.s. ($\alpha = 0.05$)} \\ 
	          Concreteness    & -0.2     & {$p<0.002$} \\
	         
	    \end{tabular}
	    \caption{Results from multiple regression that regresses rate of change in moral relevance against the factors of word frequency, length, and concreteness ($n=606$).}
	    \label{tab:regression}
	\end{table}
	
	We further verified the diachronic component of this effect in a random permutation analysis. We generated 1,000 control time courses by randomly shuffling the 20 decades in our data, and repeated the regression analysis to obtain a control distribution for each regression coefficient. All effects became non-significant under the shuffled condition, suggesting the relevance of concept concreteness for diachronic change in moral sentiment (see Supplementary Material).

\section{Discussion and conclusion}

We presented a text-based framework for exploring the socio-scientific problem of moral sentiment change. Our methodology uses minimal parameters and exploits implicit moral biases learned from diachronic word embeddings to reveal the public's moral perception toward a large concept repertoire  over a long historical period. 

Differing from existing work in NLP that treats moral sentiment as a flat classification problem \cite{garten2016morality,johnson2018classification}, our framework probes moral sentiment change at multiple levels and captures moral dynamics concerning relevance, polarity, and fine-grained categories informed by Moral Foundations Theory \cite{graham2013moral}. We applied our methodology to the automated analyses of moral change both in individual concepts and at a broad scale, thus providing insights into psycholinguistic variables that associate with rates of moral change in the public.

Our current work focuses on exploring moral sentiment change in English-speaking cultures. Future research should evaluate the appropriateness of the framework to probing moral change from a diverse range of cultures and linguistic backgrounds, and the extent to which moral sentiment change interacts and crisscrosses with linguistic meaning change and lexical coinage.  Our work creates opportunities for applying natural language processing toward characterizing moral sentiment change in society.

\section*{Acknowledgments}

We would like to thank Nina Wang, Nicola Lacerata, Dan Jurafsky, Paul Bloom, Dzmitry Bahdanau, and the Computational Linguistics Group at the University of Toronto for helpful discussion. We would also like to thank Ben Prystawski for his feedback on the manuscript. JX is supported by an NSERC USRA Fellowship and YX is funded through a  SSHRC Insight Grant, an NSERC Discovery Grant, and a Connaught New Researcher Award.

%\newpage

\bibliography{emnlp-ijcnlp-2019}
\bibliographystyle{acl_natbib}

%\appendix

%\section{Supplemental Material}
%\label{appendix}

\end{document}

\maketitle 

\appendix
\beginsupplement

\renewcommand*{\thefootnote}{\fnsymbol{footnote}}
\footnotetext[1]{Equal contribution.}
\renewcommand*{\thefootnote}{\arabic{footnote}}
\setcounter{footnote}{0}

\section{Model evaluation on historical corpora}

Table~\ref{tab:models_historical} shows average seed word classification accuracy for all models on each moral classification tier for the entire 1800-1999 period. We performed this historical evaluation using Google N-grams embeddings only, since COHA did not contain seed word embeddings for all moral categories in the earliest decades. Similar to the evaluation in modern times, we observe consistent performance above chance in all models.

\begin{table}[ht]
\centering
    \begin{tabular}{lS[table-format=1.2(2)]S[table-format=1.2(2)]S[table-format=1.2(2)]}
        \toprule                                                                           
        {\bfseries Model} &  {\bfseries  Relevance}  &  {\bfseries Polarity} &  {\bfseries Category} \\                           
    \midrule                                            
        {Random} &                        0.50 &                    0.50 &                  0.10 \\       
     {Centroid} &                         0.82(1) &                 0.89(2) &     \bfseries 0.64(2) \\                            
           {N. Bayes} &         \bfseries 0.83(2) &                 0.90(1) &               0.59(2) \\                            
        {1-NN} &                          0.80(1) &                 0.89(2) &               0.57(3) \\                           
        {5-NN} &                \bfseries 0.83(1) &       \bfseries 0.91(1) &               0.62(3) \\                           
          {KDE} &               \bfseries 0.83(1) &       \bfseries 0.91(1) &    \bfseries  0.64(2) \\                            
    %\bottomrule                                                                        
\end{tabular} 
    \caption{Moral seed word classification accuracy for moral relevance, moral polarity, and fine-grained moral categories across models using Google N-grams embeddings. Mean accuracies and standard deviations across all decades in 1800--1999.}
    \label{tab:models_historical}
\end{table}

\section{Time-shuffled regression analysis}

\begin{figure}[ht]
    \centering
    \includegraphics{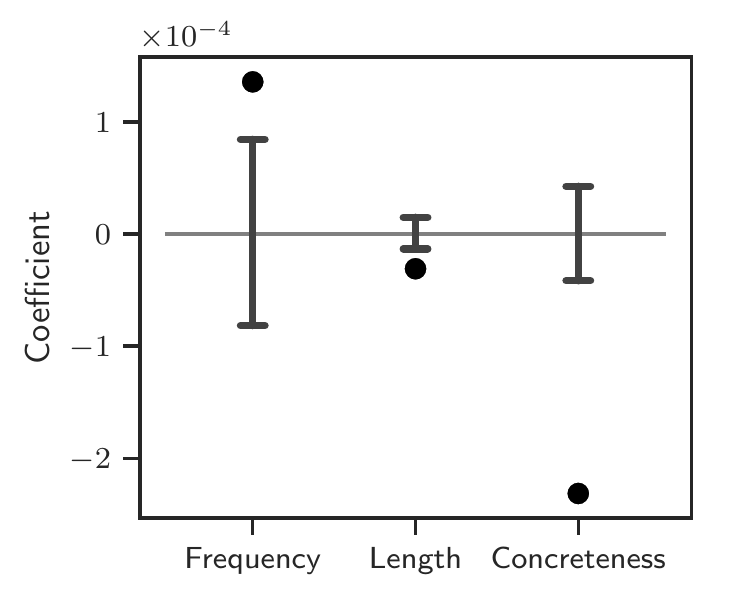}
    \caption{Multiple regression coefficients of moral relevance change versus control condition. Points show diachronic coefficients, and error bars show 1 standard deviation around the mean control coefficients.}
    \label{fig:regression}
\end{figure}

Figure~\ref{fig:regression} shows the diachronic coefficients for word frequency, length, and concreteness on moral relevance change from multiple regression analaysis, compared to the distribution of coefficients obtained from a control condition of 1,000 shuffled time courses. We observe that all effects become non-significant under the shuffled condition, and a strong diachronic effect of concreteness compared to the control suggests the relevance of this psycholinguistic variable for diachronic change in moral sentiment.

\section{Additional time courses}
\begin{figure}[!ht]
\centering
\includegraphics[width=0.495\textwidth]{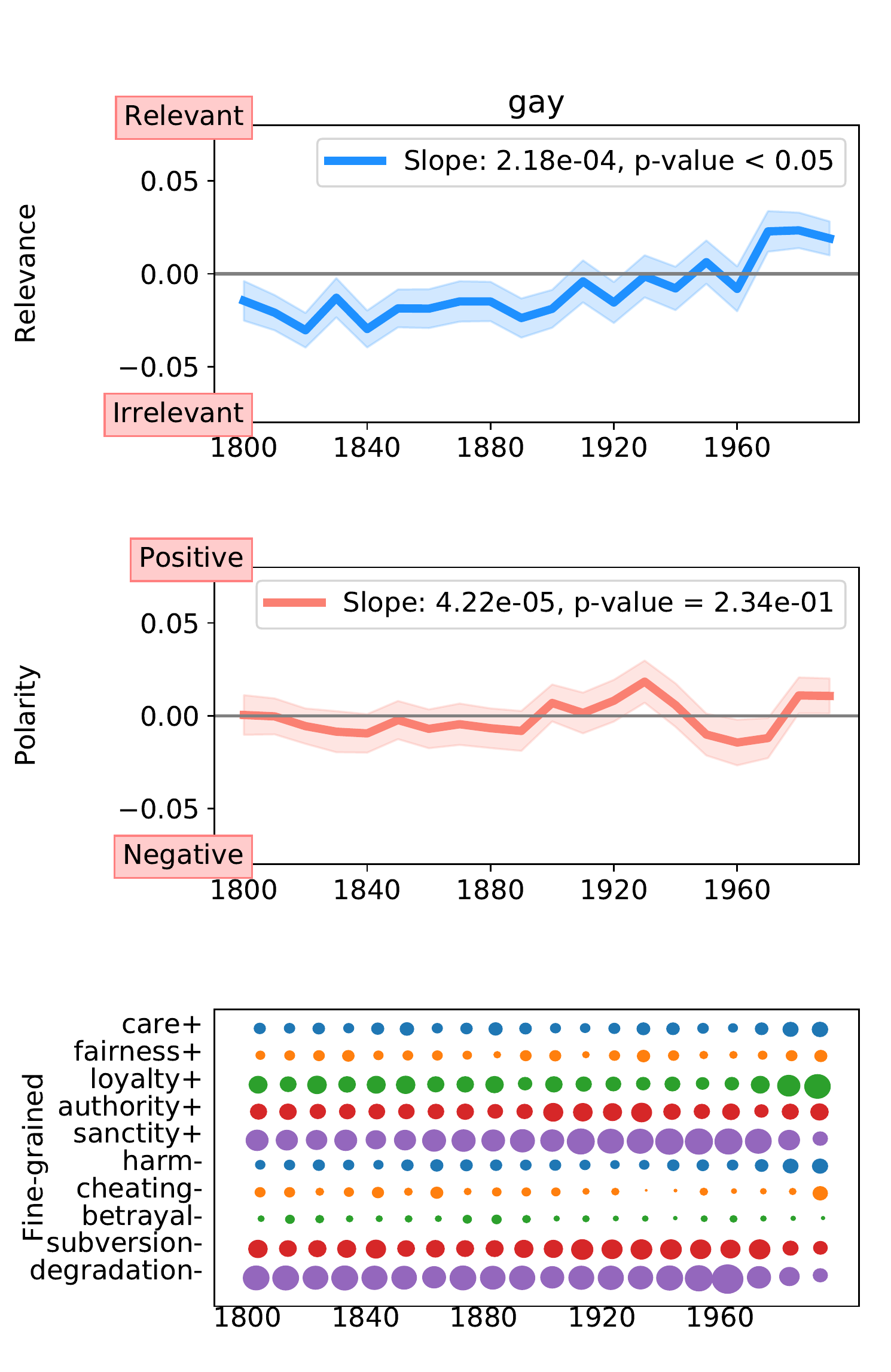}
\caption{Moral sentiment time courses of {\it gay} at each of the three levels, inferred by the Centroid model. Time courses at the moral relevance and polarity levels are in log odds ratio, and those for the fine-grained moral categories are represented by circles with sizes proportional to category probabilities.}
\label{fig:time_course_gay}
\end{figure}

Figure \ref{fig:time_course_gay} illustrates the moral trajectories for gay. Although gay originally referred to that which is ``skittish/spirited" in the 1800s, it later acquired the meaning of ``pertaining to homosexuality" in the 1930s \cite{historical_thesaurus_english}. This shift in semantics to a subject of more controversy is reflected in the moral relevance plot.

\newpage

\bibliography{emnlp-ijcnlp-2019}
\bibliographystyle{acl_natbib}